\documentclass[hyphens,sigconf,natbib=true,anonymous=false]{acmart}
\usepackage[linesnumbered,ruled,vlined]{algorithm2e}
\usepackage{todonotes}
\usepackage{lscape}
\usepackage{multirow}
\usepackage{pifont}
\usepackage{hyperref}
\usepackage{subcaption}
\newcommand{\cmark}{\ding{51}}%
\newcommand{\xmark}{\ding{55}}%
\AtBeginDocument{%
  \providecommand\BibTeX{{%
    \normalfont B\kern-0.5em{\scshape i\kern-0.25em b}\kern-0.8em\TeX}}}

\setcopyright{acmlicensed}
\copyrightyear{2022} 
\acmYear{2022} 
\acmDOI{10.1145/3477495.3531751}

\acmConference[SIGIR '22]{Proceedings of the 45th International ACM SIGIR Conference on Research and Development in Information Retrieval}{July 11--15, 2022}{Madrid, Spain}
\acmBooktitle{Proceedings of the 45th International ACM SIGIR Conference on Research and Development in Information Retrieval (SIGIR '22), July 11--15, 2022, Madrid, Spain}
\acmPrice{15.00}
\acmISBN{978-1-4503-8732-3/22/07}

\SetKwInput{KwInput}{Input}                % Set the Input
\SetKwInput{KwOutput}{Output}              % set the Output

\begin{document}

\fancyhead{}

\title[]{Knowledge Graph Question Answering Datasets and Their Generalizability: Are They Enough for Future Research?}

\author{Longquan Jiang}
\orcid{0000-0002-7333-2589}
\affiliation{%
  \institution{University Hamburg}
  \streetaddress{Mittelweg 177}
  \city{Hamburg}
  \country{Germany}}
\email{longquan.jiang@uni-hamburg.de}

\author{Ricardo Usbeck}
\orcid{0000-0002-0191-7211}
\affiliation{%
  \institution{University Hamburg}
  \streetaddress{Mittelweg 177}
  \city{Hamburg}
  \country{Germany}}
\email{ricardo.usbeck@uni-hamburg.de}


\begin{abstract}
%No. 

Existing approaches on Question Answering over Knowledge Graphs (KGQA) have weak generalizability. That is often due to the standard i.i.d. assumption on the underlying dataset. 
Recently, three levels of generalization for KGQA were defined, namely \textit{i.i.d.}, \textit{compositional}, \textit{zero-shot}.
We analyze 25 well-known KGQA datasets for 5 different Knowledge Graphs (KGs).
We show that according to this definition many existing and online available KGQA datasets are either not suited to train a generalizable KGQA system or that the datasets are based on discontinued and out-dated KGs.
Generating new datasets is a costly process and, thus, is not an alternative to smaller research groups and companies.
In this work, we propose a mitigation method for re-splitting available KGQA datasets to enable their applicability to evaluate generalization, without any cost and manual effort.
We test our hypothesis on three KGQA datasets, i.e., LC-QuAD, LC-QuAD 2.0 and QALD-9).
Experiments on re-splitted KGQA datasets demonstrate its effectiveness towards generalizability. 
The code and a unified way to access 18 available datasets is online at \url{https://github.com/semantic-systems/KGQA-datasets} as well as \url{https://github.com/semantic-systems/KGQA-datasets-generalization}.
\end{abstract}




\maketitle

\section{Introduction}
Existing approaches on Question Answering over Knowledge Graphs (KGQA) have weak generalizability since they operate with an i.i.d. assumption. The research field is missing well-defined datasets for training and evaluating systems towards generalizability.

Recently, Gu et al.~\cite{gu2021beyond}  defined three levels of generalization, namely, \textit{i.i.d.}, \textit{compositional}, \textit{zero-shot}. 
First, KGQA systems should be able to handle all elements seen in the training set, such as classes, relations, and query components as well as query structures (\textit{i.i.d.}). 
Second, the systems should be able to answer questions using known classes, relations, query components but new \textit{compositions} of these in the test set.
Third, generalizable KGQA systems should also be able to handle unseen classes, relations, and query components and structures, i.e., show \textit{zero-shot} behavior.
Consequently, Gu et al.~\cite{gu2021beyond} released a large-scale annotated KGQA dataset -- GrailQA -- for generalization evaluation. 
Note, these three levels do build on each other. That is, a test dataset which tests \textit{compositionality} does comply with \textbf{i.i.d.} behavior. 
However, GrailQA is limited to Freebase, a discontinued and outdated knowledge graph. In general, data collection and annotation for a different domain or knowledge graph is costly. An expansion to other KGs such as DBpedia, Wikidata is hard because this requires costly annotation of the logical queries to these KGs~\cite{DBLP:conf/i-semantics/GusmitaJVRNU19}.

The rise of the transformer-architecture~\cite{DBLP:conf/nips/VaswaniSPUJGKP17,DBLP:journals/widm/ChakrabortyLMTL21} injected a reawakened increase in KGQA research. Our research opens a new evaluation perspective on KGQA systems and thus, fosters and guides the development of new KGQA systems and datasets.

To cost-efficiently reuse datasets from nearly a decade of research and thus generate more training and evaluation coverage, we proposed a novel method for re-splitting KGQA datasets according to the three generalization levels without any cost.  

To answer whether existing datasets are suitable for non-iid generalization evaluation, we analyze 25 well-known datasets. For 18 of these datasets that are online available, we establish a unified way to access them. We analyze the level of generalizability in each of these. 
To show the importance of well-split datasets, we train three recent KGQA systems (TeBaQA~\cite{bfraunhofer2021knowledge}, BART~\cite{lewis-etal-2020-bart}, and HGNet~\cite{chen2021outlining}) on the original and on re-splitted datasets. Our experiments expose the weaknesses of the original datasets and shine a light on the generalization ability of the KGQA systems.

Our contributions are as follows:
\begin{itemize}
    \item Analysis of existing KGQA datasets and systems capability to generalize.
    \item Development of a novel method for cost-effective re-splitting datasets to effectively train the generalization ability of KGQA systems. 
    \item A library for loading the KG datasets \url{https://github.com/semantic-systems/KGQA-datasets}. The library is based on the Hugging Face datasets library and uses an Apache 2.0 license.
    \item We also provide the code for generating generalizable versions of these datasets using an Apache 2.0 license.
\end{itemize}

As partners of the German national research data infrastructure, we will update and maintain the repositories for at least five more years.

%\todo[inline]{Is the resource well documented? What level of expertise do you expect is required to make use of the resource? Are there tutorials or examples? Do they resemble actual uses or are they toy examples? If the resource is data, are appropriate tools provided for loading that data?If the resource is data, are the provenance (source, preprocessing, cleaning, aggregation) stages clearly documented? => @Longquan: Describe in the README of https://github.com/semantic-systems/KGQA-datasets how to load a dataset with a python example and link to the original papers not only a list. Also, show in https://github.com/semantic-systems/KGQA-datasets-generalization how to run the generalization script.} 

\section{Notation}
\subsection{Knowledge Graphs}
Here, a knowledge graph is a multigraph, denoted by $\mathcal{KG}=(V, E)$, where $V$ is a set of vertices and $E \subseteq (\mathcal{E} \cup \mathcal{B}\cup \mathcal{C}) \times \mathcal{R} \times (\mathcal{E} \cup \mathcal{B} \cup \mathcal{C} \cup \mathcal{L})$ is a set of edges expressed as RDF triples. With $\mathcal{R}$ being the set of relations, $\mathcal{E}$ being the set of entities, $\mathcal{B}$ a set of blank nodes indicating the existence of a resource with uniquely identifying it, $\mathcal{C}$ being the set of classes and $\mathcal{L}$ being the set of non-unique literal values such as strings and dates. The knowledge graph can be further subdivided into the ontology, commonly referred to the schema or T-Box $\mathcal{O} \subseteq \mathcal{C} \times \mathcal{R} \times \mathcal{C}$ and the set of facts, instances or so-called A-Box $M \subseteq (\mathcal{E} \cup \mathcal{B}) \times \mathcal{R} \times (\mathcal{E}\cup \mathcal{B} \cup \mathcal{C} \cup \mathcal{L})$~\cite{gu2021beyond}.

\subsection{Knowledge Graph Question Answering}

KGQA aims at retrieving the correct answer or respectively answers denoted as $A$ of a given natural language question $Q_{NL}$ from the knowledge graph $\mathcal{KG}$. This can be done by learning a transformation function $f$ translating the question $Q_{NL}$ into a semantically equivalent logical form $Q_{LF}$, i.e., $\displaystyle f : Q_{NL} \mapsto Q_{LF}$. In our case, we use SPARQL as the logical form. A SPARQL Abstract Query is a tuple $(A, DS, F)$, where $A$ is a SPARQL algebra expression, $DS$ is an RDF knowledge graph and $F$ is a logical query form with operators $\displaystyle\odot$ such as SPARQL solution modifiers, FILTER expressions or operators.\footnote{\url{https://www.w3.org/TR/rdf-sparql-query/\#sparqlDefinition}}

Note, there are also graph traversal-based KGQA systems which are out-of-scope of this work. 

\subsection{Generalizability in KGQA}
The definition of the three levels of generalization in KGQA stems from the work by Gu et al.~\cite{gu2021beyond}. We clarify it as follows for better contextualization. 

Let $S=\mathcal{R}\cup\mathcal{C}\cup \displaystyle\odot$ be the set of \textit{schema items}.
Let $S_{train}\subseteq S$ be the set of \textit{schema items} in the training split of a dataset. Let  $S_q\subseteq S$ be the set of \textit{schema items} in the question at hand. The test set $Q$ is defined as follows:

\begin{itemize}
    \item  \textit{I.I.D. generalization:} $\forall q \in Q: S_q \subset S_{train}$ and $q$ follows the training distribution. That is, all relations $\mathcal{R}$, classes $\mathcal{C}$ and logical form constructs $\displaystyle\odot$ have been seen while training but not the actual entities $\mathcal{E}$ and literals $\mathcal{L}$. In particular, the specific logical form constructs are known.
    \item \textit{Compositional generalization:} This level differs from \textit{I.I.D. generalization} in that the specific logical form construct and its operators $\displaystyle\odot$ are not part of $S_{train}$ but relations and classes are all known.
    \item \textit{Zero-shot generalization:} $\forall q \in Q: \exists s \in S_q, s\in S \setminus S_{train}$ That is, for all test questions, there is at least one schema item $\mathcal{R}\cup\mathcal{C}\cup \displaystyle\odot$ per question which is not in the training dataset.
\end{itemize}

To reiterate, the levels do build on each other. However, it is important to build not only one but three different test sets for evaluation. 

\section{Related Work}
Our work is in line with research on KGQA benchmarking and system evaluation. A wide range of KGQA benchmarks and datasets as well as analyses thereof have been created to evaluate KGQA systems for simple and complex questions over different publicly available Knowledge Graphs (KGs). This includes datasets such as WebQuestions~\cite{berant2013semantic} and GrailQA~\cite{gu2021beyond} for Freebase, LC-QuAD~\cite{trivedi2017lc} and QALD-9~\cite{DBLP:conf/semweb/UsbeckGN018} for DBpedia, RuBQ~\cite{korablinov2020rubq} and CronQuestions~\cite{saxena2021question} for Wikidata. However, none of them are geared towards generalization in KGQA. GrailQA is an exception, but its process of large-scale manual annotation is highly expensive and it is targeted at the not-maintained Freebase. QALD-9 is the only other dataset with similar distributional characteristics as GrailQA, but it is very small with only 558 question-answer pairs.

Bouziane et al.~\cite{bouziane2015question} surveyed a total of 31 different KGQA systems with respect to specific characteristics, e.g., interfaces to databases, ontologies, etc.  They presented an analysis of the quality of these KGQA systems using multiple indicators, e.g., success rate, correct answers.
Saleem et al.~\cite{saleem2017question} conducted a fine-grained analysis of the state-of-the-art KGQA systems on QALD-6~\cite{DBLP:conf/esws/UngerNC16}, and demonstrated that various natural language phenomena and SPARQL features would affect their final performance.  
Affolter et al.~\cite{affolter2019comparative} presented another survey of 24 KGQA systems, but they focused on general KGQA systems, not on KGQA systems specifically.
Steinmetz and Sattler~\cite{steinmetz2021kgqa} analysed into 26 popular KGQA datasets (e.g., QALD, LC-QuAD), their structure and characteristics like ambiguity, lexical gap, complex queries, etc. Their experimental results revealed the specific challenges existing KGQA systems are facing and are required to overcome. 
Although there are a lot of overview articles looking into KGQA systems and their performance, none of these looks deeply into the generalization aspect of datasets and, thus, their ability to help train generalizable KGQA systems.

To enhance evaluation insights, there are approaches modifying one particular dataset to evaluate specific characteristics of KGQA systems.
Siciliani et al.~\cite{siciliani2021mqald} presented the MQALD dataset to evaluate how well KGQA systems can work with certain SPARQL modifiers.
They extracted queries with modifiers from QALD and added manually created queries to form a new dataset.
To bridge the gap between a user and semantic data in the e-commerce industry, Kaffee et al.~\cite{DBLP:conf/kcap/KaffeeESV19} extended the base dataset (QALD-9~\cite{DBLP:conf/semweb/UsbeckGN018}) by adding SPARQL queries over different KGs. The new dataset is dubbed QALM\footnote{QALM is available at \url{https://github.com/luciekaffee/QALM}}.
Singh et al.~\cite{singh2020no} as well as Ding et al.~\cite{ding-etal-2019-leveraging} re-split LC-QuAD~\cite{trivedi2017lc} to use only queries answerable over an updated KG, i.e., DBpedia 2018 instead of 2016 to overcome outdated queries resulting in 3,253 question-answer pairs instead of 5,000.

We are the first to systematically offer a re-spliting mechanism to enable better training for generalizability in KGQA.

\section{Generation of Splits}
\subsection{Generalizability}
Existing KGQA systems have weak generalizability due to the i.i.d assumption they operate with. That is, they assume that any test set follows the distribution of seen characteristics in a given training dataset.  In addition, the shortage of evaluation sets for the ability to generalize hinders the further development of KGQA systems.

We show that the definition of generalization levels is not clear enough and contradicting. That is, the paper~\cite{gu2021beyond} suggests that the level of generalization depend on each other - compare Fig 1~\cite{gu2021beyond} - but on the contrary, they do not depend on each other. We measure whether the provided dataset GrailQA follows its definition. We demonstrate that it does not. Thus, we clarify the definition of~\cite{gu2021beyond} and provide methods to check for the different levels of generalization. We also establish a method to re-split any dataset to follow the provided definition. These novel datasets allows for more challenging KGQA research.

\subsection{Test for three levels of generalization}~\label{sec:test}
Given a KGQA dataset, to know whether and how much it is suitable for generalization evaluation of KGQA systems, we developed a simple algorithm to test which of the definitions holds for its test split. Algorithm~\ref{alg:determinelevel} describes the way of determining the level of generalization of a given question $q$ w.r.t. a given training set. Note, $d_{S}$ is a dictionary of unique, sorted tuples of schema terms, which allows us to distinguish whether this composition has already been seen in the training dataset. Each of the sets of sorted schema terms becomes a  key in the dictionary in the form of a tuple. The value is the index or id of this tuple of schema terms, set to the current length of the dictionary $d_S$. For example: \\

\noindent\textit{How many countries are on the continent of North America?} \\ 

\noindent which has the corresponding SPARQL query over Wikidata KG: \\ 

\noindent\begin{verbatim}
SELECT (COUNT(DISTINCT ?country) AS ?result) WHERE 
{
?country wdt:P31 wd:Q6256. 
?country wdt:P30 wd:Q49.
}
\end{verbatim}
\noindent which has results in the following dictionary entry: \\ 

\noindent\begin{verbatim}
Dict()-{("COUNT", "wdt:P30", "wdt:P31",): 0}
\end{verbatim}

\begin{algorithm}[!htb]
\DontPrintSemicolon
    \KwInput{A given question $q$, the set of unique schema terms in training set $S_{train}$, a dictionary of unique sets of schema terms $d_{S}$}
    \KwOutput{The level of generalization of one given question $l_q$}

    %Get the set $S_q$ of schema terms $\mathcal{R}\cup\mathcal{C}\cup \displaystyle\odot$ of a given question $q$

   % \tcc{Get the set of schema terms of $q$ unseen in the full set of schema terms}
    
    %\hfill
    
    $S_{unseen}$ $\leftarrow$ $S_q \setminus S_{train}$
    
    %\hfill
    
   % \tcc{Determine the level of the question $q$}
    
    \textbf{if} $S_q \notin d_{S}$ \textbf{then}
        
        \quad \textbf{if}{$S_{unseen } \ne \emptyset$} \textbf{then} $l_q$ $\leftarrow$ \textit{"zero-shot"}
        
        \quad \textbf{else} $l_q$ $\leftarrow$ \textit{"compositional"}
    
    \textbf{else} $l_q$ $\leftarrow$ \textit{"iid"}
    
    \Return $l_q$

\caption{Determine the level of generalization of one given question \textit{determine\_level}.}
\label{alg:determinelevel}
\end{algorithm}

\subsection{Generation Method}
To overcome the limited number of available training datasets towards generalization, we implemented a method to generate new splits for a given dataset. That is, we can generate three different subsets for \textit{i.i.d.}, \textit{compositional}, and \textit{zero-shot} generalization. We maintain the same proportions of train/test splits as the original dataset as far as possible.

Assume that we have a KGQA dataset $D=\{(q, l, s_q) | q \in Q, l \in LQ, s_q \subset S\}$. We define $q$ as a given natural language question, $Q$ is the set of all questions, $l$ is the logic form, e.g., SPARQL, of the question $q$, $LQ$ is the set of all possible logic forms, $s_q$ is the set of schema terms extracted from $l$ using pattern matching, $S$ is the set of unique schema terms of all questions $Q$. For schema terms, we only consider relations, classes, COUNT, and comparative functions (e.g., "<", ">", "<=", ">=", and "!="), similar to~\cite{gu2021beyond}. In the following, we use the complete dataset to produce the new train and test splits.

In our approach, we have two variables (see Algorithm \ref{alg:determinelevel}): (1) $d_s$, a set of unique schema terms for all questions $Q$ in the dataset $D$, and (2) $d_S$, a dictionary to store unique tuples of schema terms for all the questions $Q$ in the dataset $D$. Each of the sets of sorted schema terms becomes a  key in the dictionary in the form of a tuple (see Section~\ref{sec:test}). The value is the index or id of this tuple of schema terms. First, we assign a group id \textit{gid} to each question $q$ in the dataset $D$ by getting the key of the set of schema terms of $q$ in $d_S$. Second, we start to re-split the given dataset. The goal is to re-split a given dataset $D$ into two subsets $D_{train}$ and $D_{test}$, where $D_{test}$ have further three subsets, i.e., $D_{zero}$, $D_{compo}$, and $D_{iid}$ w.r.t. three levels of generalization respectively. See Algorithm~\ref{alg:help1} and Algorithm~\ref{alg:help2} in the appendix for more details. In the end, the generalizability of a KGQA model can be evaluated on these new splits.

Note, our algorithm allows to set the size of each of these subsets ($D_{zero}$, $D_{compo}$, $D_{iid}$), i.e., one can emphasize one particular generalization level over the others.

\begin{algorithm}[!htb]
\DontPrintSemicolon
    \KwInput{The initial KGQA dataset $D_0$; sampling ratios $R_{zero}$, $R_{compo}$, and $R_{iid}$ w.r.t. three levels; random state \textit{seed}}
    \KwOutput{
    $D_{zero}$, $D_{compo}$, $D_{iid}$, $D_{train}$
    }
    
    \tcc{\textbf{Step 1: sampling for zero-shot level}}
    
    \quad $D_1^*$, $D_{zero}^*$ = do\_GSS($D_0$, $R_{zero}$, \textit{seed})
    
    \quad $\tilde{D}_{zero}$, $\tilde{D}_{compo}$, $\tilde{D}_{iid}$ = do\_splitting($D_{zero}^*$, $D_1^*$)
    
    \quad $D_1$ = concatenate($D_1^*$, $\tilde{D}_{compo}$, $\tilde{D}_{iid}$)
    
    \hfill
    
    \tcc{\textbf{Step 2: sampling for compositional level}}
    
    \quad $D_2^*$, $D_{compo}^*$ = do\_GSS($D_1$, $R_{compo}$, \textit{seed})
    
    \quad $\tilde{D}_{zero}$, $\tilde{D}_{compo}$, $\tilde{D}_{iid}$ = do\_splitting($D_{compo}^*$, $D_2^*$)
    
    \quad $D_2$ = concatenate($D_2^*$, $\tilde{D}_{zero}$, $\tilde{D}_{iid}$)
    
    \hfill
    
    \tcc{\textbf{Step 3: sampling for iid level}}
    
    \quad $D_{train}$, $\tilde{D}_{iid}$ = train\_test\_split($D_2$, $R_{iid}$, \textit{seed})
    
    \hfill
    
    \tcc{\textbf{Step 4: final splitting}}
    
    \quad $\tilde{D}_{test}$ = concatenate($\tilde{D}_{zero}$, $\tilde{D}_{compo}$, $\tilde{D}_{iid}$)
    
    \quad $D_{zero}$, $D_{compo}$, $D_{iid}$ = do\_splitting($\tilde{D}_{test}$, $D_{train}$)
    
    \hfill
    
    \quad \textbf{Return} $D_{zero}$, $D_{compo}$, $D_{iid}$, $D_{train}$

\caption{Dataset Re-splitting Algorithm.}
\label{alg:resplit}
\end{algorithm}

Algorithm~\ref{alg:resplit} describes our overall procedure for re-splitting a given KGQA dataset. First, we sample candidate questions for the corresponding level of generalization in a descending way, from the highest level (\textit{zero-shot}) to the lowest level (\textit{i.i.d.}). The candidate question set previously sampled for a higher level will not affect lower-level sets. Note that, in each sampling round, an additional filtering process, i.e., \textit{do\_splitting} and \textit{concatenation}, is required, so that in the next round we ensure that the candidate question set drawn from the remaining full set excludes the previously sampled questions. %Moreover, according to the definitions of three levels of generalization by Gu et al., these three splits are mutually exclusive. 
In Algorithm \ref{alg:resplit}, $D_{i}$, $i \in \{1, 2\}$ represents the resultant data set as the full training set for the next sampling step. After the final sampling step, i.e., Step 3, the resultant data set $D_{train}$ would be the final training set for the given KGQA dataset $D_0$ using our approach re-splitting approach. For the two helper functions, i.e., \textit{do\_GSS} and \textit{do\_splitting}, see their implementations in Section \ref{sec:helperfunctions} of Appendix. For the \textit{train\_test\_split} function, we use the official implementation\footnote{\url{https://scikit-learn.org/stable/modules/generated/sklearn.model\_selection.train\_test\_split.html}} of the sklearn library.

\section{Analysis}

\begin{table*}[!htb]
\caption{Evaluation results of 25 common KGQA datasets analyzed in this work w.r.t. three levels of generalization. \cmark indicates that the test dataset has questions from this level of generalization. \xmark~~means there is not a single question for this level of generalization. (\cmark) or (\xmark) indicates that we deem it suitable or not suitable through literature analysis. The dash symbol '-' indicates that the level of generalization could neither be measured by literature nor by dataset analysis due to missing SPARQL queries ($*$) or valid train/test splits ($\clubsuit$).}
\begin{tabular}{cccccc}
\toprule
\multirow{2}{*}{\textbf{Dataset}} & \multirow{2}{*}{\textbf{KG}} & \multirow{2}{*}{\textbf{Year}} & \multicolumn{3}{c}{\textbf{Generalization}}                   \\ \cline{4-6}
   &     &                                & \textbf{I.I.D.} & \textbf{Compositional} & \textbf{Zero-Shot} \\ 
   \midrule
WebQuestions~\cite{berant2013semantic}                    & Freebase   & 2013 & (\cmark) & (\xmark) & (\xmark) \\ 
SimpleQuestions~\cite{bordes2015large}                 & Freebase   & 2015 & (\cmark) & (\xmark) & (\xmark) \\ 
ComplexQuestions~\cite{bao2016constraint}                & Freebase   & 2016 & $-^*$ & $-^*$ & $-^*$ \\ 
GraphQuestions~\cite{su2016generating}                   & Freebase   & 2016 & (\cmark) & (\cmark) & (\xmark) \\  
WebQuestionsSP~\cite{yih2016value}                  & Freebase   & 2016 & (\cmark) & (\xmark) & (\xmark) \\ 
The 30M Factoid QA~\cite{serban2016generating}              & Freebase   & 2016 & \cmark & \xmark & \xmark \\ 
SimpleQuestionsWikidata~\cite{DBLP:conf/esws/DiefenbachSM17}         & Wikidata   & 2017 & \cmark & \xmark & \xmark \\ 
LC-QuAD 1.0~\cite{trivedi2017lc}                      & DBpedia    & 2017 & \cmark & \cmark & \cmark \\ 
ComplexWebQuestions~\cite{talmor2018web}             & Freebase   & 2018 & (\cmark) & (\xmark) & (\xmark) \\ 
QALD-9~\cite{DBLP:conf/semweb/UsbeckGN018}                        & DBpedia    & 2018 & \cmark & \cmark & \cmark \\ 
PathQuestion~\cite{zhou2018interpretable}                   & Freebase   & 2018 & $-^\clubsuit$ & $-^\clubsuit$ & $-^\clubsuit$ \\ 
MetaQA~\cite{zhang2018variational}                          & WikiMovies & 2018 & $-^*$ & $-^*$ & $-^*$ \\ 
SimpleDBpediaQA~\cite{azmy2018farewell}                 & DBpedia    & 2018 & \cmark & \xmark & \xmark \\ 
TempQuestions~\cite{jia2018tempquestions}                  & Freebase   & 2018 & $-^\clubsuit$ & $-^\clubsuit$ & $-^\clubsuit$ \\ 
LC-QuAD 2.0~\cite{dubey2019lc}                     & Wikidata   & 2019 & \cmark & \cmark & \cmark \\ 
FreebaseQA~\cite{jiang2019freebaseqa}                      & Freebase   & 2019 & $-^*$ & $-^*$ & $-^*$ \\ 
Compositional Freebase Questions~\cite{keysers2019measuring} & Freebase   & 2020 & \cmark & \cmark & \xmark \\ 
RuBQ~\cite{korablinov2020rubq}                            & Wikidata   & 2020 & $-^\clubsuit$ & $-^\clubsuit$ & $-^\clubsuit$ \\ 
GrailQA~\cite{gu2021beyond}                         & Freebase   & 2020 & (\cmark) & (\cmark) & (\cmark) \\ 
Event-QA~\cite{souza2020event}                        & EventKG    & 2020 & $-^\clubsuit$ & $-^\clubsuit$ & $-^\clubsuit$ \\ 
RuBQ 2.0~\cite{rybin2021rubq}                        & Wikidata   & 2021 & $-^\clubsuit$ & $-^\clubsuit$ & $-^\clubsuit$ \\ 
MLPQ\footnote{\url{https://github.com/tan92hl/MLPQ}}                            & DBpedia    & 2021 & $-^*$ & $-^*$ & $-^*$ \\ 
Compositional Wikidata Questions~\cite{cui2021multilingual} & Wikidata   & 2021 & \cmark & \cmark & \xmark \\ 
TimeQuestions~\cite{jia2021complex}                   & Wikidata   & 2021 & $-^\clubsuit$ & $-^\clubsuit$ & $-^\clubsuit$ \\ 
CronQuestions~\cite{saxena2021question}                    & Wikidata   & 2021 & $-^*$ & $-^*$ & $-^*$ \\ 
 \bottomrule
\end{tabular}
\label{tab:level_results_existing_qa_system}
\end{table*}

\subsection{Experimental Setup}

Our main evaluation goal is to thoroughly evaluate the effectiveness of dataset re-splitting in increasing the generalization ability of KGQA systems. We use the following three KGQA models to verify our hypothesis. 

\begin{itemize}
    \item TeBaQA~\cite{bfraunhofer2021knowledge} is a template-based KGQA system, which constructs the correct SPARQL query by first mapping natural language questions into isomorphic basic graph pattern classes and then filling semantic information. Since TeBaQA has very high runtimes, we use it only on smaller datasets.
    \item  BART~\cite{lewis-etal-2020-bart} is a sequence-to-sequence model based on a denoising autoencoder, which is fine-tuned on the machine translation task. We use BART to evaluate performance on LC-QuAD and LC-QuAD 2.0. In particular on Wikidata-based datasets, due to its speed and original purpose.
    \item HGNet~\cite{chen2021outlining} is a hierarchical autoregressive generation model, which completes the query graph by first outlining its structure and then filling the instances into these structures. HGNet is used to evaluate different DBpedia datasets as originally intended by the authors.
\end{itemize}

\paragraph{Evaluation Metrics} For measuring the performance, we adopt Precision, Recall, F1 score, and Hits@1.

\paragraph{Implementation Details} Our experiments are twofold: (1) we perform a  preliminary analysis on the original datasets, (2) we evaluate the re-splitted datasets. For both stages, the model parameters are fixed and consistent with the original papers. For the HGNet model, we use the same configurations as in~\cite{chen2021outlining}. To be specific, we set the size of hidden layers to 256, the number of BiLSTM layers to 1, the number of the graph transformer layer to 3, the learning rate to $2 \times 10^{-4}$, the size of each beam to 7. For the BART model, we also use the same hyperparameters as in~\cite{chen2021outlining}. For the TeBaQA system \footnote{see \url{https://github.com/dice-group/TeBaQA}}, we deploy a local instance on our server and test the three KGQA datasets using the default query API.

\paragraph{Knowledge Graphs}. In this work, we use DBpedia 2016-10\footnote{\url{https://downloads.dbpedia.org/2016-10/}} for evaluating all KGQA systems on LC-QuAD and QALD-9. We host a local SPARQL query endpoint via Virtuoso to retrieve answers. In addition, we also test the BART system on LC-QuAD 2.0 against the live version of Wikidata. For querying the Wikidata KG, we use the official live endpoint \footnote{\url{https://query.wikidata.org/sparql}}.

\subsection{Test existing KGQA Datasets}

\begin{figure*}[!tb]
\begin{subfigure}{.33\textwidth}
  \centering
  % include first image
  \includegraphics[width=\linewidth]{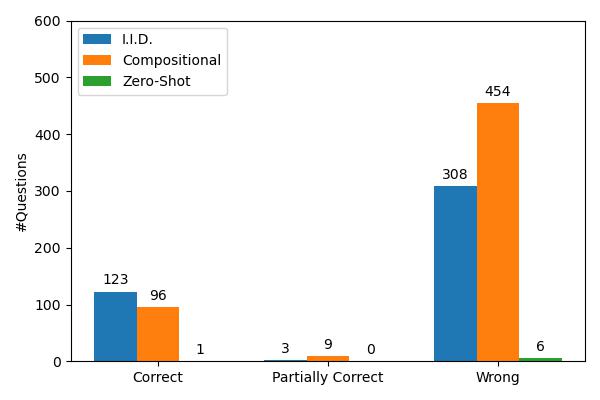}
  \caption{BART trained on LC-QuAD.}
  \label{fig:baseline_lcquad_preliminary_analysis}
\end{subfigure}
\begin{subfigure}{.33\textwidth}
  \centering
  % include first image
  \includegraphics[width=\linewidth]{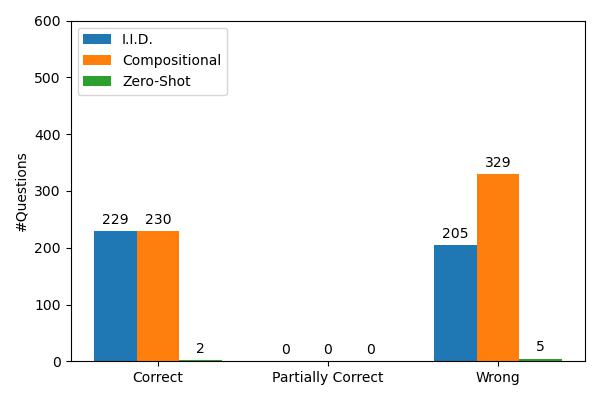}
  \caption{HGNet trained on LC-QuAD.}
  \label{fig:hgnet_lcquad_preliminary_analysis}
\end{subfigure}
\begin{subfigure}{.33\textwidth}
  \centering
  % include second image
  \includegraphics[width=\linewidth]{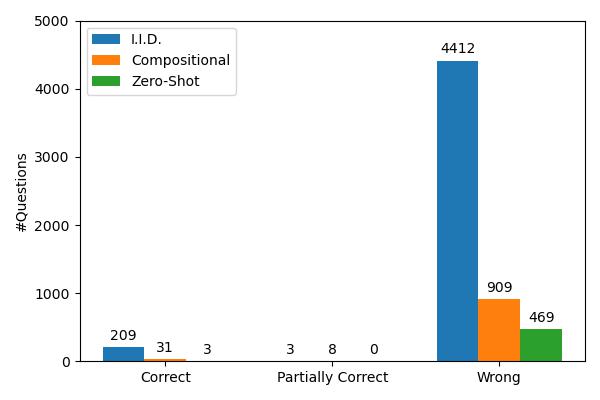}
  \caption{BART trained on LC-QuAD 2.0.}
  \label{fig:baseline_lcquad2_preliminary_analysis}
\end{subfigure}
\caption{The preliminary analytical results of the re-splitted datasets.}
\label{fig:preliminary_analysis}
\end{figure*}

To facilitate the development of KGQA systems in the Semantic Web community, researchers have generated many KGQA datasets over different publicly available Knowledge Graphs for specific purposes. We examined 25 popular KGQA datasets (and their splits) towards their inherent ability to train the generalization ability of KGQA systems, see Table~\ref{tab:level_results_existing_qa_system}. We also calculated the percent of questions w.r.t. three levels for 8 dataset which provide SPARQL queries, see Table~\ref{tab:level_results_existing_qa_system_2} in the appendix. Note, a \cmark does not automatically mean, the dataset is suitable for the evaluation but indicates the sole existence of queries from this level of generalization. From this evaluation, we have the following observations:

\begin{enumerate}
    \item All KGQA datasets targeting Freebase, except for GrailQA, lack the capability to evaluate zero-shot generalization.
    \item Out of the KGQA datasets targeting DBpedia or Wikidata, QALD-9, LC-QuAD 1.0 and LC-QuAD 2.0 are inherently able to evaluate these three levels of generalization, but they suffer from unbalanced or small-scale issues.
    \item Most KGQA datasets have no capability to evaluate compositional generalization. The exceptions are CFQ and CWQ which are specifically designed for measuring the composition ability of KGQA systems.
    \item Simple factoid KGQA datasets (e.g., SimpleQuestions, SimpleDBpediaQA, etc.) can only be used to evaluate i.i.d. generalization. This is due to the structure of simple questions involving only one basic graph pattern and thus having no complex structure.
\end{enumerate}

Additionally, we conducted a preliminary performance analysis of the BART and HGNet KGQA systems using LC-QuAD and LC-QuAD 2.0. These systems use neural networks and thus should have a better generalization ability than template-based approaches such as TeBaQA. We calculate the number of questions answered correctly, partially correct, and incorrect w.r.t. different levels of generalization. Our analytical results are shown in Figure~\ref{fig:preliminary_analysis}. %Figure~\ref{fig:baseline_lcquad_preliminary_analysis}, Figure~\ref{fig:baseline_lcquad2_preliminary_analysis}, and Figure~\ref{fig:hgnet_lcquad_preliminary_analysis}. 
We find that i.i.d. and compositional questions contribute a lot to the final performance of BART and HGNet on LC-QuAD, as they account for the largest proportion in LC-QuAD. A similar observation holds for BART on LC-QuAD 2.0. HGNet was developed for DBpedia and Freebase, not Wikidata and thus was not evaluated.

The results verify our assumption: the tested KGQA datasets are not sufficient to train KGQA systems for higher levels of generalizability. Note, we directly use the original splits to train and test KGQA systems. One solution would be to construct a brand-new specific KGQA dataset from scratch for generalization evaluation. However, the process of data collection and annotation would be prohibitively expensive and inflexible. Moreover, we think existing KGQA datasets can and should be recycled if we can design an algorithm to re-split them.

\subsection{Re-splitting existing Datasets}

We hypothesize that existing KGQA datasets could be also applicable to evaluate generalization if we re-split them. We test our hypothesis on three popular KGQA datasets, i.e., LC-QuAD, LC-QuAD 2.0, and QALD. LC-QuAD*, LC-QuAD 2.0*, and QALD* are derived from their original versions using our proposed re-splitting methodology.

\paragraph{QALD}
%need to rewrite that in a more formal way
%In section 5.2 we demonstrate that QALD-9 dataset is not a good one for machine learning systems. Indeed, we find that performance does not improve if we “expand” the training set, but it does improve if we add to train questions *similar in structure* to the test set. Basically, the training set is not representative.
%Again, we want to remark that to the expansion made to the QALD9 dataset we do not add the test set data in the training data, but we generate a new dataset from the test set patterns (By changing the entities) and include it in the training set, as explained in section 5.2 third paragraph. Indeed we specify that "No pair or gold query from the original test set was added."
In this work, we use QALD-9\cite{DBLP:conf/semweb/UsbeckGN018}, the most recent QALD challenge at the time of writing. It consists of 508 multilingual questions in total which are compiled and curated from the previous challenges and expert annotators. In particular, the test set is always created from scratch for each challenge. Note, its small size hinders adoption in end-to-end large-scale KGQA systems. 

\paragraph{LC-QuAD}
This KGQA dataset is the first large-scale complex question dataset over DBpedia, containing 4000 training and 1000 test question-answer pairs. The answers are provided as SPARQL queries. Note, the questions and queries are based on templates~\cite{trivedi2017lc}. 

\paragraph{LC-QuAD 2.0}
In comparison to LC-QuAD, the updated version - LC-QuAD 2.0~\cite{dubey2019lc} - has 30,000 unique pairs of questions and SPARQL queries, with a higher language variation and more question types. LC-QuAD 2.0 provides DBpedia and Wikidata SPARQL queries. However, these are automatically generated so  their quality can not be guaranteed. Thus, here we only use the provided Wikidata SPARQL queries, from which schema terms are extracted.

\begin{table}[htb!]
    \centering
    \caption{Percent of questions with different levels of generalization in the given test sets of QALD, LC-QuAD, and LC-QuAD 2.0.}
    \begin{tabular}{cccc}
    \toprule
    & \textbf{I.I.D} & \textbf{Compositional} & \textbf{Zero-Shot} \\
    \midrule
      QALD & 30.67\% & 35.33\% & 34.00\% \\
      LC-QuAD & 43.40\% & 55.90\% & 0.70\% \\
      LC-QuAD 2.0 & 76.51\% & 15.68\% & 7.81\% \\
      \bottomrule
    \end{tabular}
    \label{tab:dataset_overview}
\end{table}

\begin{table}[htb!]
    \centering
    \caption{Percent of questions with different levels of generalization in the given test sets of QALD*, LC-QuAD*, and LC-QuAD 2.0*.}
    \begin{tabular}{cccc}
    \toprule
             & \textbf{I.I.D} & \textbf{Compositional} & \textbf{Zero-Shot} \\
    \midrule
      QALD* & 8.09\% & 23.70\% & 68.21\% \\
      LC-QuAD* & 20.95\% & 64.62\% & 14.43\% \\
      LC-QuAD 2.0* & 40.55\% & 32.68\% & 26.78\% \\
    \bottomrule
    \end{tabular}
    \label{tab:dataset_overview_new}
\end{table}

% \begin{figure}[htb!]
%     \centering
%     \includegraphics[width=\linewidth]{images/dataset_overview.jpg}
%     \caption{Percent of questions with different levels of generalization in the given test sets of QALD, LC-QuAD, and LC-QuAD 2.0.}
%     \label{fig:dataset_overview}
% \end{figure}
% \begin{figure}[htb!]
%     \centering
%     \includegraphics[width=\linewidth]{images/dataset_overview_new.jpg}
%     \caption{Percent of questions with different levels of generalization in the given test sets of QALD*, LC-QuAD*, and LC-QuAD 2.0*.}
%     \label{fig:dataset_overview_new}
% \end{figure}

\begin{table}[htb!]
\caption{Dataset Statistics.}
\resizebox{\linewidth}{!}{
\begin{tabular}{cccccccc}
\toprule
\multirow{2}{*}{\textbf{Dataset}} &
  \multirow{2}{*}{\textbf{Total}} &
  \multirow{2}{*}{\textbf{Train}} &
  \multirow{2}{*}{\textbf{Dev}} &
  \multicolumn{4}{c}{\textbf{Test}} \\ \cline{5-8} 
 &
   &
   &
   &
  \multicolumn{1}{c}{\textbf{Total}} &
  \multicolumn{1}{c}{\textbf{I.I.D.}} &
  \multicolumn{1}{c}{\textbf{Compositional}} &
  \textbf{Zero-Shot} \\ \midrule
QALD-9        & 558   & 408   & -    & \multicolumn{1}{c}{150}  & \multicolumn{1}{c}{46}   & \multicolumn{1}{c}{53}   & 51   \\ 
QALD-9 *      & 558   & 385   & -    & \multicolumn{1}{c}{173}  & \multicolumn{1}{c}{14}   & \multicolumn{1}{c}{41}   & 118  \\ 
LC-QuAD       & 5000  & 4000  & -    & \multicolumn{1}{c}{1000} & \multicolumn{1}{c}{434}  & \multicolumn{1}{c}{559}  & 7    \\ 
LC-QuAD *     & 5000  & 3420  & 521  & \multicolumn{1}{c}{1059} & \multicolumn{1}{c}{331}  & \multicolumn{1}{c}{1021} & 228  \\ 
LC-QuAD 2.0   & 30221 & 20321 & -    & \multicolumn{1}{c}{6633} & \multicolumn{1}{c}{4624} & \multicolumn{1}{c}{948}  & 472  \\ 
LC-QuAD 2.0* & 30221 & 24177 & 3267 & \multicolumn{1}{c}{6044} & \multicolumn{1}{c}{4014} & \multicolumn{1}{c}{3235} & 2651 \\ 
\bottomrule
\end{tabular}}
\label{tab:dataset_statistics}
\end{table}

\iffalse
\begin{figure*}
\begin{subfigure}{.33\textwidth}
  \centering
  % include first image
  \includegraphics[width=\linewidth]{images/baseline_lcquad_new_preliminary_analysis.jpg}
  \caption{BART trained on LC-QuAD*.}
  \label{fig:baseline_lcquad_new_preliminary_analysis}
\end{subfigure}
\begin{subfigure}{.33\textwidth}
  \centering
  % include second image
  \includegraphics[width=\linewidth]{images/baseline_lcquad2_new_preliminary_analysis.jpg}
  \caption{BART trained on LC-QuAD 2.0*.}
  \label{fig:baseline_lcquad2_new_preliminary_analysis}
\end{subfigure}
\begin{subfigure}{.33\textwidth}
  \centering
  % include first image
  \includegraphics[width=\linewidth]{images/hgnet_lcquad_new_preliminary_analysis.jpg}
  \caption{HGNet trained on LC-QuAD*.}
  \label{fig:hgnet_lcquad_new_preliminary_analysis}
\end{subfigure}
\caption{Analytical results of the re-splitted datasets.}
\label{fig:new_preliminary_analysis}
\end{figure*}
\fi

\begin{figure*}
\begin{subfigure}{.45\textwidth}
  \centering
  % include first image
  \includegraphics[width=\linewidth]{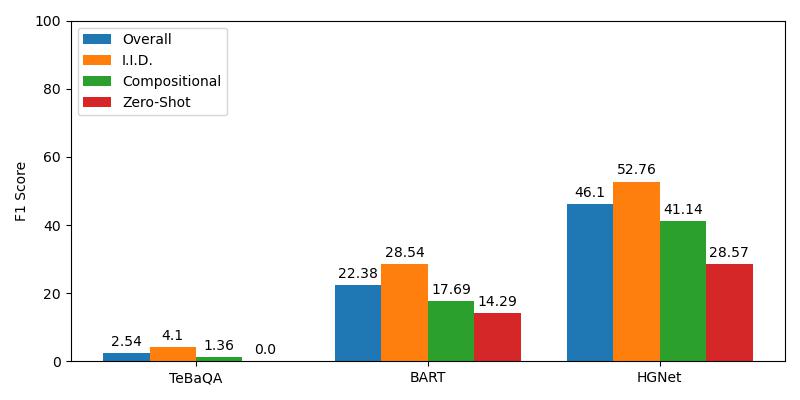}
  \caption{Evaluation of all systems on LC-QuAD.}
  \label{fig:result_lcquad}
\end{subfigure}
\begin{subfigure}{.45\textwidth}
  \centering
  % include first image
  \includegraphics[width=\linewidth]{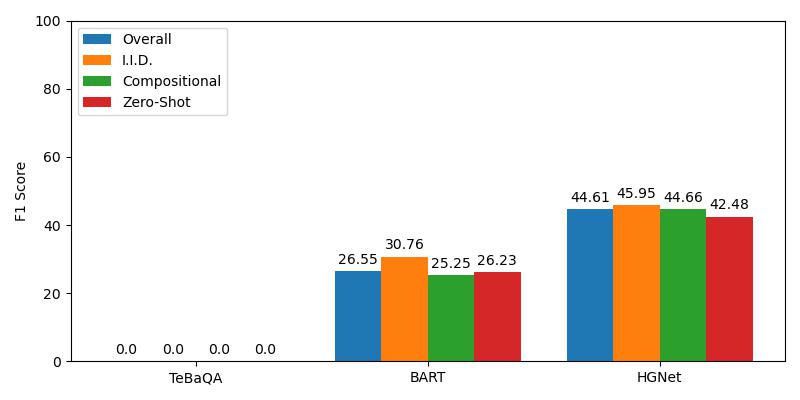}
  \caption{Evaluation of all systems on LC-QuAD*.}
  \label{fig:result_lcquad_new}
\end{subfigure}
\begin{subfigure}{.45\textwidth}
  \centering
  % include first image
  \includegraphics[width=\linewidth]{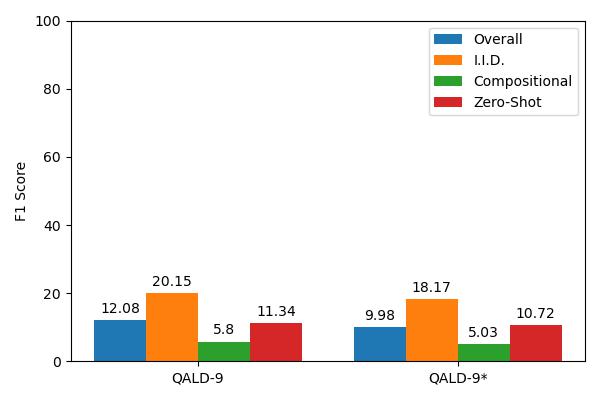}
  \caption{TeBaQA trained and tested on QALD and QALD*.}
  \label{fig:result_tebaqa_qald}
\end{subfigure}
\begin{subfigure}{.45\textwidth}
  \centering
  % include second image
  \includegraphics[width=\linewidth]{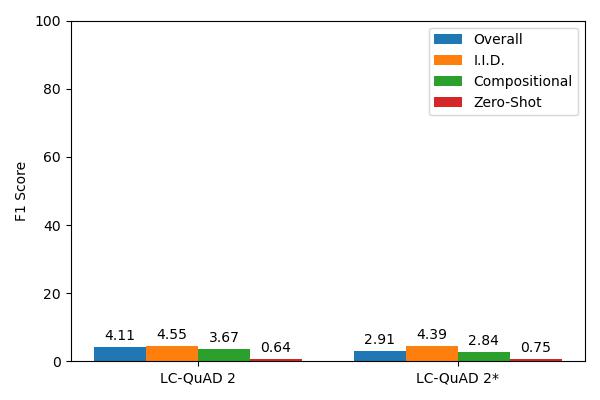}
  \caption{BART trained and tested on LC-QuAD 2.0 and LC-QuAD 2.0*.}
  \label{fig:result_lcquad2}
\end{subfigure}
\caption{The evaluation results of the re-splitted datasets.}
\label{fig:evaluation_results}
\end{figure*}

Table~\ref{tab:dataset_overview} shows the statistics of these three datasets w.r.t. different levels of generalization without re-splitting. We find that questions for stronger generalization levels account for a smaller percentage of the test set, except for QALD-9, which has a balanced distribution. 
Note, since the original QALD-9 dataset is suited for i.i.d evaluation, we increased the \textit{zero-shot percentage disproportionately on purpose}. 
LC-QuAD has the most unbalanced distribution. There are only 7 questions for measuring zero-shot generalization, accounting for 0.7\% - much smaller compared to the other two levels. This unbalanced distribution constraints the applicability of LC-QuAD to evaluate stronger generalization of KGQA systems trained with it despite its scale. Note, LC-QuAD 2.0 is more balanced.

Validation of KGQA systems for generalizability depends on the splits on which they are trained and tested. Thus, we employ our re-splitting algorithm. The statistics of re-splitted datasets are shown in Table~\ref{tab:dataset_overview_new}. Compared to the original splits of these three datasets, the new datasets emphasize balancing question distributions w.r.t. three levels of generalization, especially for the zero-shot level. Table~\ref{tab:dataset_statistics} shows the details of original and re-splitted versions of these three KGQA datasets.

\subsection{Evaluating the Re-split}
To assess the performance boost due to the use of re-splitted datasets in training, we run experiments on re-splitted versions of QALD, LC-QuAD, and LC-QuAD 2.0 datasets. The evaluation results are shown in Figure~\ref{fig:evaluation_results}.

In detail, Figure~\ref{fig:result_lcquad} and Figure~\ref{fig:result_lcquad_new} show the F1 scores obtained by testing TeBaQA, BART and HGNet on LC-QuAD and LC-QuAD* respectively. 

We find that the F1 scores of TeBaQA on the re-splitted version of LC-QuAD w.r.t. different levels decrease to 0.0, indicating that the TeBaQA system does not have the capability of being generalizable within LC-QuAD. The main reason is that in the original split some test question features leaked into the train dataset due to the template-based design of the dataset. Thus, this result does not indicate generalizability. That is, due to the complex and diverse structure of questions, the SPARQL templates which TeBaQa learns to map a question cannot be learned with a proper split. % so it is impossible to cover much more questions for different levels of generalization. 
Figure~\ref{fig:result_tebaqa_qald} shows the F1 scores using TeBaQA on QALD and its re-splitted version. Compared with the original splits, QALD* poses a significant challenge to the final performance of TeBaQA. 

To show the scalability of our approach, we also tested the BART system on LC-QuAD 2.0 which is based on the large-scale Wikidata KG. The results indicate, our approach can also ensure the good quality and difficulty of the new splits on large datasets and on a different KG, namely, Wikidata while increasing the proportions of compositional and zero-shot questions (see~Figure \ref{fig:result_lcquad2}). BART performance worsens on LC-QuAD 2.0* as opposed to its performance on LC-QuAD 2. This is due to the new balance, where before LC-QuAD was biased towards i.i.d. and compositional questions, and its re-split is more evenly distributed. Thus, the evaluation shows an artifact. LC-QuAD 2.0 was less biased and is now overall more balanced, demanding more generalizability and, thus, the system performs worse. 

Overall, one can observe a more realistic evaluation of KGQA systems towards their capability to generalize to unseen schema items and logical forms.

%\todo[inline]{Error Analysis?}
%Systems are better but not good enough, why? What are they missing? Are particular %SPARQL structures just not frequent enough in the trianing sets?

\section{Summary and Future Work}
Research on the ability to generalize for any artificial intelligence system, such as KGQA systems, is important and requires high-quality datasets.
We introduced a test for three levels of generalization capability evaluation over KGQA datasets.
We developed a method to re-split existing KGQA datasets to avoid unfair evaluations as well as costly generation of new datasets. 
For the first time, our evaluation spots a gap in the KGQA evaluation landscape pointing towards test leakage. Using re-split datasets, we show that the investigated systems are not able to generalize well on re-split datasets (QALD*, LC-QuAD 2.0*) which do not have test leakage. Additionally, for newly balanced datasets (LC-QuAD), we show that performance can go up.

In the future, we will extend our generation methods for more detailed evaluations which in turn can be used by agent-based KGQA systems to train themselves on weak points. 
We will also integrate this evaluation paradigm in our benchmarking platform~\cite{DBLP:journals/semweb/UsbeckRHCHNDU19} to sustain its use.

We hope to advance the understanding of the generalization ability of different KGQA systems and to support the KGQA community. 

\begin{acks}
The authors acknowledge the financial support by the Federal Ministry for Economic Affairs and Climate Action of Germany in the project CoyPu (project number 01MK21007G).
\end{acks}
\bibliographystyle{ACM-Reference-Format}
\bibliography{sample-base}


\appendix

\section{Helper Functions}{\label{sec:helperfunctions}}

To implement our re-splitting algorithm, we also designed the following two helper functions. Algorithm \ref{alg:help1} aims to offer randomized train/test indices to split a given dataset according to a third-party provided group. We used the official GroupShuffleSplit implementation\footnote{\url{https://scikit-learn.org/stable/modules/generated/sklearn.model\_selection.GroupShuffleSplit.html}} of the scikit-learn library\footnote{\url{https://scikit-learn.org}}. For each sampling step of three levels, Algorithm \ref{alg:help1} return train/test splits from the remaining dataset (e.g., $D_0$, $D_1$, $D_2$) in the previous step. The \textit{test} split would be the candidate subset for the corresponding sampling step, while the \textit{train} split would be the full set for the next sampling step. Algorithm \ref{alg:help2} aims to split a test data $D_{test}$ into three subsets w.r.t. three levels of generalization, i.e., $D_{zero}$, $D_{compo}$, and $D_{iid}$, according to the training set $D_{train}$.

\begin{algorithm}[htb!]
\DontPrintSemicolon
    \KwInput{KGQA dataset $D$; sampling ratio $ratio$; random state \textit{seed}}
    \KwOutput{
    train/test splits \textit{train} and \textit{test}
    }
    
    \hfill
    
    \SetKwFunction{DoGSS}{do\_GSS}
    \SetKwFunction{DoSplitting}{do\_splitting}
    
    \SetKwProg{Fn}{Function}{:}{}
  \Fn{\DoGSS{\textit{D}, \textit{ratio}, \textit{seed}}}{
        Get group ids \textit{gids} of sets of the questions in \textit{D} \;
        gss = GroupShuffleSplit(1-\textit{ratio},\textit{seed})\;
        \textit{train}, \textit{test} = gss.split(X=\textit{D}, groups=\textit{gids})\;
        \KwRet \textit{train}, \textit{test}\;
  }
  \;
\caption{Group Shuffle Split.}
\label{alg:help1}
\end{algorithm}

\begin{algorithm}[htb!]
\DontPrintSemicolon
    \KwInput{Testing set $D_{test}$; Training set $D_{train}$}
    \KwOutput{
    $D_{zero}$, $D_{compo}$, $D_{iid}$
    }
    
      \SetKwProg{Fn}{Function}{:}{}
  \Fn{\DoSplitting{$D_{test}$, $D_{train}$}}{
        Get the set of schema terms $d_s$ for all questions in $D_{train}$ \;
        Get the dictionary of unique sets of schema terms $d_S$ for all questions in $D_{train}$ \;
        
        Define three lists of questions $D_{zero}$, $D_{compo}$, and $D_{iid}$ for three levels \;
        
        \textbf{foreach} $q \in$ $D_{test}$ \textbf{do}
        
        \quad $l_q$ = determine\_level($d_s$, $d_S$, $q$)
        
        \quad \textbf{if} $l_q ==$ \textit{"zero-shot"} \textbf{then} append $q$ to $D_{zero}$
        
        \quad \textbf{elif} $l_q ==$ \textit{"compositional"} \textbf{then} append $q$ to $D_{compo}$
        
        \quad \textbf{elif} $l_q ==$ \textit{"iid"} \textbf{then} append $q$ to $D_{iid}$
        
        \textbf{end for}
         
        \KwRet $D_{zero}$, $D_{compo}$, $D_{iid}$\;
  }
  \;
\caption{Generalization Level Split.}
\label{alg:help2}
\end{algorithm}

%\begin{landscape}
\begin{table*}[htb!]
\caption{Evaluation Results on LC-QuAD, QALD, LC-QuAD 2.0 and their corresponding re-splits.}
\resizebox{\linewidth}{!}{
\begin{tabular}{cccccccccccccccccc}
\toprule
\multirow{2}{*}{} &
  \multirow{2}{*}{\textbf{Model}} &
  \multicolumn{4}{c}{\textbf{Overall}} &
  \multicolumn{4}{c}{\textbf{I.I.D.}} &
  \multicolumn{4}{c}{\textbf{Compositional}} &
  \multicolumn{4}{c}{\textbf{Zero-Shot}} \\ \cline{3-18} 
 & &
  \multicolumn{1}{c}{\textbf{P}} &
  \multicolumn{1}{c}{\textbf{R}} &
  \multicolumn{1}{c}{\textbf{F1}} &
  \textbf{Hits@1} &
  \multicolumn{1}{c}{\textbf{P}} &
  \multicolumn{1}{c}{\textbf{R}} &
  \multicolumn{1}{c}{\textbf{F1}} &
  \textbf{Hits@1} &
  \multicolumn{1}{c}{\textbf{P}} &
  \multicolumn{1}{c}{\textbf{R}} &
  \multicolumn{1}{c}{\textbf{F1}} &
  \textbf{Hits@1} &
  \multicolumn{1}{c}{\textbf{P}} &
  \multicolumn{1}{c}{\textbf{R}} &
  \multicolumn{1}{c}{\textbf{F1}} &
  \textbf{Hits@1} \\ 
  \midrule
\multirow{3}{*}{LC-QuAD *} &
  HGNet &
  \multicolumn{1}{c}{49.48} &
  \multicolumn{1}{c}{49.48} &
  \multicolumn{1}{c}{49.48} &
  49.48 &
  \multicolumn{1}{c}{50} &
  \multicolumn{1}{c}{50} &
  \multicolumn{1}{c}{50} &
  50 &
  \multicolumn{1}{c}{49.35} &
  \multicolumn{1}{c}{49.35} &
  \multicolumn{1}{c}{49.35} &
  49.35 &
  \multicolumn{1}{c}{49.24} &
  \multicolumn{1}{c}{49.24} &
  \multicolumn{1}{c}{49.24} &
  49.24 \\ 
 &
  BART &
  \multicolumn{1}{c}{26.58} &
  \multicolumn{1}{c}{27.36} &
  \multicolumn{1}{c}{26.55} &
  26.72 &
  \multicolumn{1}{c}{31.21} &
  \multicolumn{1}{c}{30.99} &
  \multicolumn{1}{c}{30.76} &
  31.53 &
  \multicolumn{1}{c}{25.17} &
  \multicolumn{1}{c}{26.24} &
  \multicolumn{1}{c}{25.25} &
  25.29 &
  \multicolumn{1}{c}{26.19} &
  \multicolumn{1}{c}{27.12} &
  \multicolumn{1}{c}{26.23} &
  26.14 \\ 
 &
  TeBaQA &
  \multicolumn{1}{c}{0} &
  \multicolumn{1}{c}{0} &
  \multicolumn{1}{c}{0} &
  0 &
  \multicolumn{1}{c}{0} &
  \multicolumn{1}{c}{0} &
  \multicolumn{1}{c}{0} &
  0 &
  \multicolumn{1}{c}{0} &
  \multicolumn{1}{c}{0} &
  \multicolumn{1}{c}{0} &
  0 &
  \multicolumn{1}{c}{0} &
  \multicolumn{1}{c}{0} &
  \multicolumn{1}{c}{0} &
  0 \\ 
  \midrule
\multirow{3}{*}{LC-QuAD} &
  HGNet &
  \multicolumn{1}{c}{52.14} &
  \multicolumn{1}{c}{52.14} &
  \multicolumn{1}{c}{52.14} &
  52.14 &
  \multicolumn{1}{c}{58.42} &
  \multicolumn{1}{c}{58.42} &
  \multicolumn{1}{c}{58.42} &
  58.42 &
  \multicolumn{1}{c}{47.24} &
  \multicolumn{1}{c}{47.24} &
  \multicolumn{1}{c}{47.24} &
  47.24 &
  \multicolumn{1}{c}{40} &
  \multicolumn{1}{c}{40} &
  \multicolumn{1}{c}{40} &
  40 \\ 
 &
  BART &
  \multicolumn{1}{c}{22.45} &
  \multicolumn{1}{c}{22.94} &
  \multicolumn{1}{c}{22.38} &
  22.4 &
  \multicolumn{1}{c}{28.5} &
  \multicolumn{1}{c}{28.85} &
  \multicolumn{1}{c}{28.54} &
  28.34 &
  \multicolumn{1}{c}{17.86} &
  \multicolumn{1}{c}{18.47} &
  \multicolumn{1}{c}{17.69} &
  17.89 &
  \multicolumn{1}{c}{14.29} &
  \multicolumn{1}{c}{14.29} &
  \multicolumn{1}{c}{14.29} &
  14.29 \\ 
 &
  TeBaQA &
  \multicolumn{1}{c}{2.72} &
  \multicolumn{1}{c}{2.61} &
  \multicolumn{1}{c}{2.54} &
  2.7 &
  \multicolumn{1}{c}{4.16} &
  \multicolumn{1}{c}{4.1} &
  \multicolumn{1}{c}{4.1} &
  4.14 &
  \multicolumn{1}{c}{1.63} &
  \multicolumn{1}{c}{1.48} &
  \multicolumn{1}{c}{1.36} &
  1.61 &
  \multicolumn{1}{c}{0} &
  \multicolumn{1}{c}{0} &
  \multicolumn{1}{c}{0} &
  0 \\ 
  \midrule
QALD-9* &
  TeBaQA &
  \multicolumn{1}{c}{10.8} &
  \multicolumn{1}{c}{12.54} &
  \multicolumn{1}{c}{9.98} &
  9.83 &
  \multicolumn{1}{c}{20.86} &
  \multicolumn{1}{c}{17.05} &
  \multicolumn{1}{c}{18.17} &
  21.43 &
  \multicolumn{1}{c}{4.24} &
  \multicolumn{1}{c}{12.18} &
  \multicolumn{1}{c}{5.03} &
  2.44 &
  \multicolumn{1}{c}{11.89} &
  \multicolumn{1}{c}{12.13} &
  \multicolumn{1}{c}{10.72} &
  11.02 \\ 
  \midrule
QALD-9 &
  TeBaQA &
  \multicolumn{1}{c}{13.87} &
  \multicolumn{1}{c}{12.34} &
  \multicolumn{1}{c}{12.08} &
  14 &
  \multicolumn{1}{c}{21.19} &
  \multicolumn{1}{c}{21.16} &
  \multicolumn{1}{c}{20.15} &
  19.57 &
  \multicolumn{1}{c}{8.63} &
  \multicolumn{1}{c}{6.05} &
  \multicolumn{1}{c}{5.8} &
  9.43 &
  \multicolumn{1}{c}{12.71} &
  \multicolumn{1}{c}{10.92} &
  \multicolumn{1}{c}{11.34} &
  13.73 \\ 
  \midrule
LC-QuAD 2.0* &
  BART &
  \multicolumn{1}{c}{2.91} &
  \multicolumn{1}{c}{2.92} &
  \multicolumn{1}{c}{2.91} &
  2.89 &
  \multicolumn{1}{c}{4.39} &
  \multicolumn{1}{c}{4.39} &
  \multicolumn{1}{c}{4.39} &
  4.39 &
  \multicolumn{1}{c}{2.83} &
  \multicolumn{1}{c}{2.86} &
  \multicolumn{1}{c}{2.84} &
  2.77 &
  \multicolumn{1}{c}{0.75} &
  \multicolumn{1}{c}{0.76} &
  \multicolumn{1}{c}{0.75} &
  0.79 \\ 
  \midrule
LC-QuAD 2.0 &
  BART &
  \multicolumn{1}{c}{4.11} &
  \multicolumn{1}{c}{4.17} &
  \multicolumn{1}{c}{4.11} &
  4.15 &
  \multicolumn{1}{c}{4.56} &
  \multicolumn{1}{c}{4.57} &
  \multicolumn{1}{c}{4.55} &
  4.56 &
  \multicolumn{1}{c}{3.63} &
  \multicolumn{1}{c}{3.97} &
  \multicolumn{1}{c}{3.67} &
  3.9 &
  \multicolumn{1}{c}{0.64} &
  \multicolumn{1}{c}{0.64} &
  \multicolumn{1}{c}{0.64} &
  0.64 \\ 
  \bottomrule
\end{tabular}}
\label{tab:full_evaluation_results}
\end{table*}

\begin{table*}[htb!]
\caption{Analytical results of 8 KGQA datasets about the percent of questions w.r.t. three levels of generalization.}
\begin{tabular}{ccccc}
\toprule
\multirow{2}{*}{\textbf{Dataset}} & \multirow{2}{*}{\textbf{Complex Questions?}} & \multicolumn{3}{c}{\textbf{Generalization}}  \\ \cline{3-5}
 &   & \textbf{I.I.D.} & \textbf{Compositional} & \textbf{Zero-Shot} \\ 
\midrule
The 30M Factoid QA~\cite{serban2016generating}              & No   & 100.00\% & 0.00\% & 0.00\% \\ 
SimpleQuestionsWikidata~\cite{DBLP:conf/esws/DiefenbachSM17}         & No   & 100.00\% & 0.00\% & 0.00\% \\ 
LC-QuAD 1.0~\cite{trivedi2017lc}                      & Yes    & 43.40\% & 55.90\% & 0.70\% \\ 
QALD-9~\cite{DBLP:conf/semweb/UsbeckGN018}                        & Yes   & 30.67\% & 35.33\% & 34.00\% \\ 
SimpleDBpediaQA~\cite{azmy2018farewell}                 & No   & 100.00\% & 0.00\% & 0.00\% \\ 
LC-QuAD 2.0~\cite{dubey2019lc}                     & Yes & 76.51\% & 15.68\% & 7.81\% \\ 
%Compositional Freebase Questions~\cite{keysers2019measuring} & Freebase   & 2020 & \cmark & \cmark & \xmark \\ \hline
GrailQA~\cite{gu2021beyond}                         & Yes  & 25.00\% & 25.00\% & 50.00\% \\ 
Compositional Wikidata Questions~\cite{cui2021multilingual} & Yes  & 49.15\% & 50.85\% & 0.00\% \\ 
\bottomrule
\end{tabular}
\label{tab:level_results_existing_qa_system_2}
\end{table*}

\section{Evaluation Results}
We choose Precision, Recall, F1 Score, and Hits@1 as evaluation metrics for all KGQA datasets. The full evaluation results with three popular datasets, i.e., LC-QuAD~\cite{trivedi2017lc}, QALD-9~\cite{DBLP:conf/semweb/UsbeckGN018}, LC-QuAD 2.0~\cite{dubey2019lc}, and their corresponding re-splits, are shown in Table \ref{tab:full_evaluation_results}.

\section{Dataset Analysis}
Tabel~\ref{tab:level_results_existing_qa_system_2} shows the analytical results of 8 KGQA datasets about the percent of questions w.r.t. the three levels of generalization~\cite{gu2021beyond} using our open-source testing method. The result of GrailQA was copied from the original paper~\cite{gu2021beyond}.

\end{document}